# Integrating Real-Time Analysis With The Dendritic Cell Algorithm Through Segmentation

Feng Gu, Julie Greensmith and Uwe Aickelin
School of Computer Science, University of Nottingham,
Nottingham, UK, NG8 1BB.
{fxg,jqg,uxa}@cs.nott.ac.uk

## ABSTRACT

As an immune inspired algorithm, the Dendritic Cell Algorithm (DCA) has been applied to a range of problems, particularly in the area of intrusion detection. Ideally, the intrusion detection should be performed in real-time, to continuously detect misuses as soon as they occur. Consequently, the analysis process performed by an intrusion detection system must operate in real-time or near-to real-time. The analysis process of the DCA is currently performed offline, therefore to improve the algorithm's performance we suggest the development of a real-time analysis component. The initial step of the development is to apply segmentation to the DCA. This involves segmenting the current output of the DCA into slices and performing the analysis in various ways. Two segmentation approaches are introduced and tested in this paper, namely antigen based segmentation (ABS) and time based segmentation (TBS). The results of the corresponding experiments suggest that applying segmentation produces different and significantly better results in some cases, when compared to the standard DCA without segmentation. Therefore, we conclude that the segmentation is applicable to the DCA for the purpose of real-time analysis.

## Categories and Subject Descriptors

I.2 [Artificial Intelligence]: Miscellaneous

## 1. INTRODUCTION

Artificial Immune Systems (AIS) [4] are computer systems inspired by both theoretical immunology and observed immune functions, principles and models, which can be applied to real world problems. As the natural immune system is designed to protect the body from a wealth of invading micro-organisms, artificial immune systems are developed to provide the same defensive properties within a computing context. Initially AIS based themselves upon simple models of the human immune system. As noted by Stibor et al. [16], 'first generation algorithms' including negative and clonal selection do not produce the same high performance as the human immune system. These algorithms, negative selection in particular, are prone to problems with scaling and the generation of excessive false alarms when used to detect intruders in computer networks. Recently developed AIS use more rigourous and up-to-date immunology and are developed in collaboration with immunologists. The resulting algorithms are believed to encapsulate the desirable properties of immune systems including robustness and error tolerance.

One of such 'second generation' AIS is the Dendritic Cell Algorithm (DCA) [7]. This algorithm is inspired by the function of the dendritic cells of the innate immune system and incorporates the principles of a key novel theory in immunology, termed the danger theory [13]. This theory suggests that dendritic cells (DCs) are responsible for the initial detection of invading micro-organisms. An abstract model of natural DC behaviour is used as the foundation of the developed algorithm. Currently, the DCA has been successfully applied to numerous problem domains, including port scan detection [7], Botnet detection [1] and as a classifier for robotic security [14]. These applications have suggested that the DCA shows not only good performance on detection rate, but also the ability to reduce the rate of false alarms in comparison to other systems including Self Organising Maps [10]. The majority of applications to which the DCA is applied involve the detection of unauthorised use and abuse of computer systems and networks - a prob- lem termed intrusion detection. Systems designed to detect intrusions are termed intrusion detection systems. The de- velopment of reliable and sophisticated intrusion detection systems is non trivial, as such systems need to process huge amounts of data in a short period of time and simultaneously achieve high levels of detection accuracy.

As stated by Zhang et al. [17], in practice, intrusion detection is a real-time critical mission. This means that intrusions should be detected as soon as possible or at least before an attack eventually succeds. The detection speed which reflects the time taken for detecting intrusions, is the actual key to prevent successful attacks. We believe that an effec-tive intrusion detection system should ideally be a real-time system that can react to the input within the certain time bounds. The time bounds constrain the maximum latency for the system to identify to an intrusion after its appearance. Consequently, the analysis of an intrusion detection system should be done in a fast and continuous manner, namely, in real-time or near-to real-time.

The DCA can internally process input data in real-time. However, this algorithm requires a further analysis process which is thus far performed offline. To develop the DCA into a fully functioning intrusion detection system, it is desirable to improve the real-time capability of the algorithm by modifying the analysis component. Initially, it is important to decide at which point the analysis component should process its current batch of data, in order to derive intrusion scores for the identification of intrusions. This can be achieved by applying segmentation to the analysis process of the DCA. As the word suggests, segmentation involves slicing the output data into smaller segments with a view of generating finer grained results, as well as performing analysis in parallel with the detection process. Segmentation is performed based on a fixed quantity of output data items or alterna- tively on a basis of a fixed time period. Thus, segmentation enables the system to perform periodic analysis whenever sufficient information is presented during detection.

The aim of this paper is to investigate two segmentation approaches and to explore the applicability of segmentation to the DCA. The investigation is focussed on the comparison between the standard DCA without segmentation and the two newly introduced segmentation approaches. A range of segment sizes in both data quantity and time are varied to demonstrate any potential effects. To test our hypotheses we use a large real-world dataset based on a medium scale port-scan of a university computer network. We intend to use this investigation as a basis for the further work on the development of a dynamic real-time solution for the analysis of the DCA. The presented experiments are necessary steps towards achieving this aim. The paper is organised as follows: the details of the DCA are described in Section 2; real-time analysis and segmentation are demonstrated in Section 3; the experiments are explained in Section 4; the results and the analysis are reported in Section 5; and finally the conclusions and future work are drawn in Section 6.

## 2. THE DENDRITIC CELL ALGORITHM

### 2.1 The Biological Background

The DCA is a population based algorithm, capable of processing multiple input sources, originally designed to solve problems within intrusion detection. As previously stated the blueprint for the DCA is the function of the dendritic cells (DCs) of the innate immune system, which is the body's first line of defence against invaders. In nature DCs have the ability to combine a multitude of molecular information and to interpret this information for the T-cells of the adaptive immune system, to induce appropriate immune response towards perceived threats.

Signal and antigen are the two types of molecular information processed by DCs. Signals are collected by DCs from their local environment and consist of indicators of the health of the monitored tissue. DCs are sensitive to three types of signal: PAMP signals derived from molecules produced exclusively by invading micro-organisms; danger signals generated as a result of cell stress and unexpected cell death; and safe signals produced by healthy cells. In addition, DCs exist in one of three states of maturation to perform their immune function. In their initial immature state, DCs are exposed to a combination of these signals. Cells exposed to higher concentrations of PAMP and danger signal transition to a fully mature form and can instruct the adaptive immune system to activate. Conversely, higher concentration of safe signal induces partial or 'semi-maturation' of DCs and have a suppressive effect on the activation of the adaptive system.

Additionally, during their immature phase DCs also collect debris in the tissue which are subsequently combined with the molecular environmental signals. Some of the debris collected are termed antigens, and are proteins originating from potential invading entities. DCs combine the 'suspect' antigens with evidence in the form of signals to correctly instruct the adaptive immune system to respond, or become tolerant to the antigens in question. For more detailed information, refer to Lutz and Schuler [13].

The resulting algorithm incorporates the state transition pathway, the environmental signal processing procedure, and the correlation between signals and antigens. In the algorithm signals are represented as real valued numbers and antigens are categorical values of the objects to be classified. The algorithm is based on a multi-agent framework, where each cell processes its own environmental signals and collects antigens. Diversity is generated within the cell population through the application of a 'migration threshold' - this value limits the number of signal instances an individual cell can process during its lifespan. This creates a variable time window effect, with different cells processing the signal and antigen input streams over a range of time periods [15]. The combination of signal/antigen correlation and the dynamics of a cell population are responsible for the detection capabilities of the DCA. In the remainder of this section we describe the algorithmic details of the DCA implementation.

### 2.2 The Deterministic DCA

In this paper we describe, implement and apply the system based on the deterministic version of DCA (dDCA) [9] for testing our hypotheses with respect to the addition of an improved analysis module. However, a third signal category is added to the input signals due to the complexity of the testing dataset. The dDCA was introduced for providing a reproducible and tractable system that is ideal for further analysis and development. The dDCA employs a population of artificial DCs, each of which has the ability to combine multiple signal sources to assess the environmental context, as well as asynchronously sample another data stream - antigen. The correlation between signals and antigens is used as the basis of identifying the intrusions contained within the input data. To accomplish this, the standard dDCA has three phases, which are system initialisation, data processing and offline analysis, as shown in Figure 1. The system initialisation phase involves generating the initial DC population. Each DC in the population is assigned with a particular migration threshold. The migration threshold of a DC is sequentially increased by a fixed number as its index number increases. As a result, the migration thresholds of the whole DC population form a uniform distribution, which creates diversity in the population.

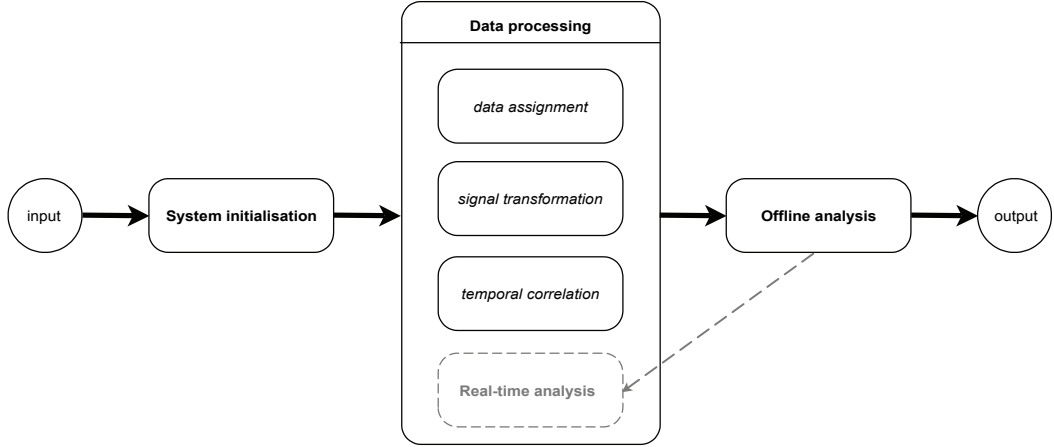

Figure 1: Three phases of the DCA and the future development of the real-time analysis, the offline analysis process will be replaced by a real-time analysis component, which performs analysis whenever sufficient information is presented during detection.

Following this phase, the input data (signals and antigens) are fed to the data processing phase. This phase consists of three sub-functions, which are data assignment, signal transformation, and temporal correlation. Firstly, the data assignment function separates signals and antigens within the input data, so that the signals are then passed to the signal transformation function, while the antigens are allocated to particular DCs selected from the population. Secondly, the signal transformation function performs the transformation from input signals to output signals. As all DCs receive the same numerical set of signals for each iteration, the processing of a single signal instance is performed globally. A user-defined number of signal sources are involved as the input signals, pre-categorised as either 'PAMP' (the third signal category added), 'danger' or 'safe'. The semantics of these signals are listed as follows:

- **PAMP**: a signature of abnormal behaviour, e.g. errors per second. An increase in this signal is associated with a high confidence of abnormality.

- **Danger**: a measure of an attribute which increases in value to indicate an abnormality. Low values of this signal may not be anomalous, giving a high value a moderate confidence of indicating abnormality.

- **Safe**: a measure which increases value in conjunction with observed normal behaviour. This is a confident indicator of normal, predictable or steady-state system behaviour. This signal is used to counteract the effects of PAMPs and danger signals.

As suggested by immunologists, safe signals always have a negative effect to output signals, and PAMP signals have a greater effect than danger signals. Such relationship is represented by predefined weights in the algorithm. Two output signals are derived from signal transformation, which are 'CSM' and 'k'. The calculation of output signals is displayed in Equation 1,

$$O_j = \sum_{i=0}^{n}(W_{ij} \times S_i) \quad \forall j \qquad (1)$$

where $O_j$ are the output signals, n is the number of output signal categories minus one, $S_i$ is the input signals and $W_{ij}$ is the transforming weight from $S_i$ to $O_j$. The output signals can be accessed by all DCs in the population. Thirdly, each DC performs a temporal correlation function between signals and antigens internally. An individual DC creates a time window specified by its migration threshold, signals and antigens which appear within this time window are correlated with each other. As suggested in [9], to perform correct correlation, the signals are supposed to appear after the antigens, and the delay should be shorter than the time window created by each DC. In the mean time, the system also performs summing of the output signals. Once a DC's cumulative CSM exceeds its migration threshold, it changes the state and becomes a matured DC. As a result it stops performing signal transformation and temporal correlation. The association between the cumulative k and sampled antigens within each DC is termed the 'processed information', which is then presented by the cell to the analysis phase. Once a matured DC has presented the processed information, it is reset to an immature DC, and hence the DC population size is kept constant.

In the offline analysis phase, all the processed information presented by the matured DCs is collated for the analysis process. The analysis is performed per 'antigen type' - a collection of identical antigen instances. The outcome of the analysis process is a measure of whether an antigen type is an intrusion or not, such measure is termed '$K_\alpha$', which is calculated by Equation 2 [9],

$$K_\alpha = \frac{\sum k_i}{\sum \alpha_i} \quad \forall i \qquad (2)$$

where $\alpha_i$ is the number of antigen type $\alpha$ sampled by DC i, and $k_i$ is its cumulative k. The greater the value of $K_\alpha$, the higher the probability that antigen type $\alpha$ is an intrusion. The dDCA implementation is shown as Algorithm 1 [9].

As mentioned previously, the analysis process in the dDCA is performed offline after the data processing phase, which is insufficiently effective for an intrusion detection system.

```
input  : antigens and signals
output: antigen types plus K_α

set DC population size;
initialize DCs;
while data do
    switch input do
        case antigen
            agCounter++;
            cellIndex = agCounter % populationSize;
            DC of cellIndex assigned antigen;
            update DC's antigen profile;
        end
        case signal
            calculate csm and k;
            foreach DC do
                DC.lifespan -= csm;
                DC.sumK += k;
                if DC.lifespan <= 0 then
                    record antigens, DC.sumK;
                    reset DC;
                end
            end
        end
    end
end
foreach antigen type do
    calculate K_α;
end
```

Algorithm 1: Pseudocode of the dDCA implementation.

Therefore, we intend to move the analysis phase to be with the data processing phase, as indicated in Figure 1. The analysis process can be then performed periodically to identify intrusions during detection, that is, in real-time.

## 3. REAL-TIME ANALYSIS

### 3.1 Real-Time Analysis And Segmentation

A real-time analysis component is essential for developing an effective intrusion detection system from the DCA. Such component performs periodic analysis of the processed information presented by DCs, to continuously identify intrusions during detection. An effective and fully functioning intrusion detection system should be able to identify the intrusions as quickly as possible, as accurately as possible, and hence detection speed and detection accuracy are two major indicators of performance. Most of the techniques can produce reasonable detection accuracy, if sufficient time is given. But as demonstrated before, detection speed is the actual key to the performance of an intrusion detection system. If an intrusion detection system fails to identify the intrusions in time, no further responses against the intrusions can be made. This leads to the eventual success of attacks, which is a fatal failure of an intrusion detection system. Therefore, if the intrusions are identified too late, even with 100% detection accuracy, it all becomes meaningless in terms of system defence. As a result, we propose integrating real-time analysis with the DCA, to improve detection speed without compromising detection accuracy.

If the real-time analysis is to be performed during detection, one issue needs resolved, that is, when the analysis should be performed. This issue could be solved by applying segmentation to the DCA. It is different from the moving time windows method described in [11], which is used in the pre-processing stage to smooth noisy input signals, as segmentation is performed in the post-processing stage for the purpose of analysis. As the processed information is presented by matured DCs over time, a sequence of processed information is being generated during detection. Sesegmentation involves partitioning this sequence into relative smaller segments, in terms of the number of data items or time. All the generated segments have the same size, and the analysis is performed within each individual segment. Therefore, in each segment, one set of detection result ($K_\alpha$ per antigen type) is generated, in which intrusions appeared within the duration of this segment can be identified.

First of all, segmentation can produce multiple sets of results, rather than one set of results produced by non-segmentation system. This enables the system to perform analysis in real-time (online), rather than offline, as all segments are processed during detection. In addition, segmentation distributes the analysis process into multiple steps, instead of performing at once. This can reduce the computation power and time required for the analysis process, so segmentation can effectively enhance detection speed. Moreover, as the processed information is presented by matured DCs at different time points over the duration, analysing the sequence of processed information at once ignores the temporal difference of each piece of processed information. As a result, same antigen type which causes malicious activities at one point but does nothing at another point may be classified as normal rather than an intrusion. This can be avoided by applying segmentation, as it features periodic analysis that can cope with the inherited time differences. Therefore, the system can effectively discriminate the activities which are intrusions from those which are not, and hence the detection accuracy is also improved.

The most important and in fact the only factor of segmentation is the segment size. It determines how soon the intrusions can be identified. The smaller the segment size, the sooner the intrusions can be identified, and vice versa. Moreover, the segment size may also influence the sensitivity of the final results. If the segment size is too large, the results can lose the sensitivity and thus the system loses the ability to identify true positives. However, if the segment size is too small, the results may be too sensitive, and the system can generate false positives. In this paper, we only introduce static segmentation with a fixed segment size to the system, as the effect of different segment sizes on the detection performance needs to be investigated first. Eventually a dynamic segmentation approach will be developed, in which the segment size varies based on the real-time situations during detection.

### 3.2 The Approaches To Segmentation

Two segmentation approaches are applied, namely 'antigen based segmentation' (ABS) and 'time based segmentation' (TBS). These two approaches set the segment size respectively according to two factors, which are the number of sampled antigens or the processed time. As data accumulates during detection, theses factors dictate at which point the analysis should be performed. The number of sam-pled antigens indicates the amount of potential suspects that have been identified by the system, that is, the quantity of objects to be classified. ABS creates a segment whenever the number of sampled antigens reaches the segment size, and the analysis is performed within this segment. Similar work was done in [12], in which the overall network traffic is partitioned into subsets of manageable size, and the analysis is performed within each partition. Conversely, the processed time implies the quantity of signals that have been processed, as one set of signals is updated once per iteration in the algorithm. The processed time determines the quantity of evidence that can be used for supporting classification. TBS creates a segment whenever the defined time period elapses, and the analysis is also performed within each segment. Such an approach is commonly used in real-time control of robotics, for example, to periodically compute the next steering command in motion planning to avoid collisions [6]. The concept of using segmentation with the DCA is not entirely novel, preliminary work of ABS and TBS has been performed in [10] and [15] respectively. However, the corresponding experiments took only a cursory glance at segmentation. In this paper we examine the addition of segmentation in much greater detail than in previous work.

|     | PAMP | Danger | Safe |
| --- | --- | --- | --- |
| CSM | 4 | 2 | 6 |
| k | 8 | 4 | -13 |

Table 1: Weights for signal transformation.

| Parameter | Value |
| --- | --- |
| Population size | 100 |
| Migration thresholds | $12 \times x, \ x \in [1, 100]$ |
| Segment size (ABS) | $1 \times 10^n, \ n \in \{2,3,4,5,6\}$ |
| Segment size (TBS) | $1 \times 10^n, \ n \in \{0,1,2,3\}$ |

Table 2: Experimental parameters.

Segment size is vital to the quantity of the number of sampled antigens or the processed time contained in each segment. In order to perform sensitivity analyses, a range of segment sizes are tested, to find out their effects on the algorithm's performance. Although ABS and TBS employ a fixed segment size, they also both involve dynamics of various system factors. In ABS the number of sampled antigens required for each segment is fixed, resulting in the processed time contained in each segment being variable. For example, one segment can last 10 seconds, another one with the same number of processed antigen can last over 30 seconds. Whereas, in TBS the time required for each segment is fixed, resulting in the number of sampled antigens contained in each segment being variable. For instance, one segment can have 100 processed antigens, and another one can have over 500 or even 1000 processed antigens. As a result, by investigating both segmentation approaches, different aspects of system behaviour can be explored. This can provide more insights into the algorithm, which are useful for further development of dynamic segmentation in the future.

## 4. THE EXPERIMENTS

We use an intrusion detection dataset to test the described segmentation approaches integrated with the dDCA. The systems are programmed in C with a gcc 4.0.1 compiler. All experiments are run on an Intel 2.2 GHz MacBook (OS X 10.5.5), with the statistical tests performed in R (2.8.1). The predefined weights used for signal transformation in Equation 1 are displayed in Table 1, they are the same as those used in previous work [10]. All other experimental parameters are listed in Table 2. Sensitivity analyses of various population sizes [9] have shown that 100 is an appropriate value to use. The fixed number related to the assignment of migration thresholds is set to ensure the migration thresholds of most DCs in the population are greater than the strength of a single signal instance, so that these DCs can last longer than one iteration.

### 4.1 The SYN Scan Dataset

SYN scan is an intrusion technique used by attackers for exploiting the vulnerabilities of victim machines. The SYN scan dataset was collected under the scenario that the scan is performed by an insider, who can be a legitimate user of the system performing unauthorised activities. The SYN scan dataset [7] is chosen as the input data of the system. This dataset was collected through an ssh connection, when both anomalous and normal processes are included. This dataset is large and noisy, making the problem difficult to solve. It is ideal for the purpose of testing, as the segmentation approaches are proposed to improve the DCA. The dataset consists of over 13 million antigen instances and more than 4800 sets of signals.

In [9] the authors use only danger and safe signal categories in dDCA for a simple dataset. But for the purpose of ease of analysis, in this paper we use all three signal categories, including PAMP, danger and safe. This dataset was originally used in [8], where all seven signals were used. Only the most appropriate three signals are selected, because the aim of this paper is to introduce the concept of segmentation rather than solving the problem. The PAMP signal is the number of ICMP 'destination unreachable' (DU) error messages received per second. When the closed ports of a host are scanned, a large amount of DU error messages can be generated by the firewall. The danger signal is based the ratio of TCP packets to all other packets processed by the network card of the scanning host. A burst of this ratio is not usually observed under normal conditions, which means something malicious could be happening. The safe signal is derived from the observation that during SYN scans the average network packet size reduces to a size of 40 bytes. The scans tend to send small sized packets in large quantity, big packet sizes indicate normal behaviours of the network. One set of signals is captured per second. All signals are normalised into the range within [0,100], to match the predefined weight for signal transformation. These signals are plotted in Figure 2.

The process IDs (PIDs) whenever a system call is made on the host are recorded as individual antigens, but only the antigen types with high frequency are of interest. The number of each interesting antigen type per second is plotted Figure 3. The antigen types of interest include 'Nmap', 'Firefox' and 'Pts'. Nmap is the program used for invoking and performing SYN scans to the victim machines, Pts (pseudoterminal slave) demon process is the parent of the Nmap process, and Firefox is performing web browsing throughout the recorded session. As a result, the antigen types of Nmap and Pts are considered to be anomalous, while the antigen type of Firefox is normal.

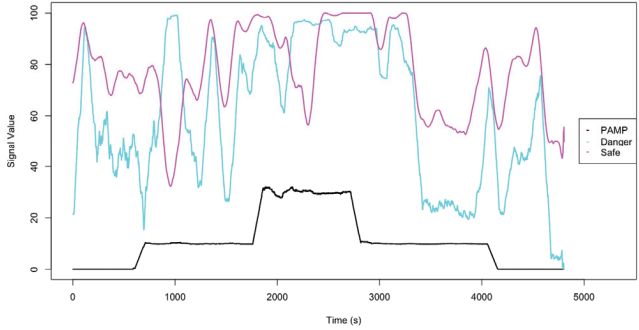

Figure 2: Input signal values against time series (moving average with intervals of 100, per selected signal category, used for plotting the graph, but not the actual input of the system).

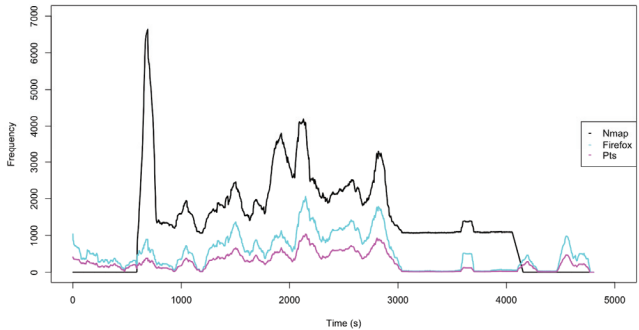

Figure 3: Number of each antigen type per second against time series (moving average with intervals of 100, per antigen type of interest, used for plotting the graph, but not the actual input of the system).

The antigens and signals are sorted according to their time stamps as the input data. As Nmap and Pts are considered responsible for intrusions, we expect to see high $K_\alpha$ values for Nmap and Pts are produced by the system during scanning activity. Whereas Firefox is considered as a normal process, so low $K_\alpha$ values for Firefox are expected.

### 4.2 Experiments And Hypotheses

Two sets of experiments are conducted, to examine the effects of ABS and TBS with the dDCA. The experiments performed are listed as follows:

1. Experiment 1 (E1): experiments using the ABS approach with various segment sizes are performed. This corresponds to the null hypothesis (H1) that changing the segment size of ABS makes no significant differences to the results. The comparisons are performed between one segment size and another.

2. Experiments 2 (E2): experiments using the TBS approach with various segment sizes are performed. This corresponds to the null hypothesis (H2) that changing the segment size of TBS makes no significant differences to the results. Similar comparisons are performed between one segment size and another.

In addition to H1 and H2, two more hypotheses can be tested by using the results of E1 and E2. The third hypothesis (H3) is that applying segmentation makes no significant difference to the results, which can be tested by the results of segmentation approaches with the result of the standard dDCA. The fourth hypothesis (H4) is that changing segmentation approach from ABS to TBS makes no significant differences to the results.

## 5. RESULTS AND ANALYSIS

### 5.1 Experimental Results

The experimental results consist of the $K_\alpha$ value per antigen type when various segment sizes are applied. One set of $K_\alpha$ values of all involved antigen types (one $K_\alpha$ value per antigen type) are generated within each segment. These $K_\alpha$ values are used for identifying the anomalous antigen types within the particular duration covered by a segment. Each row of Table 3 and Table 4 represents the statistics of all $K_\alpha$ values of an antigen type over all generated segments. The experimental results of ABS show that the minimum of $K_\alpha$ values increases, whereas the mean, maximum and standard deviation of $K_\alpha$ values decrease, as the segment size increases. Similarly, the experimental results of TBS indicate that the minimum and mean of $K_\alpha$ values increase, while the maximum and standard deviation of $K_\alpha$ values decrease, as the segment size increases. It appears that changing segment size in both segmentation approach can make differences to the results, but the differences are not obvious. As a consequence, more rigourous statistical tests need to be performed, in order to examine whether the differences are significant or not. The focus of these statistical tests is to examine the effect of segmentation with various segment sizes for each antigen type, rather than the differences between normal and anomalous antigen types as shown in [10].

### 5.2 The Statistical Tests

As all experimental results are normally distributed, a two-sample two-sided t-test ($\alpha = 0.05$) [3] is used to test

| Seg | Min | Mean | Max | Stdev |
|---|---|---|---|---|
| Nmap | | | | |
| $1 \times 10^2$ | -3358.0 | -929.9 | 679.0 | 512.8 |
| $1 \times 10^3$ | -2785.0 | -935.2 | 606.0 | 465.8 |
| $1 \times 10^4$ | -2225.0 | -934.8 | 496.6 | 372.4 |
| $1 \times 10^5$ | -1547.0 | -932.7 | 131.8 | 334.4 |
| $1 \times 10^6$ | -1157.0 | -951.5 | -354.7 | 234.1 |
| Firefox | | | | |
| $1 \times 10^2$ | -3574.0 | -963.0 | 679.0 | 539.5 |
| $1 \times 10^3$ | -3178.0 | -993.6 | 606.0 | 509.0 |
| $1 \times 10^4$ | -2806.0 | -985.5 | 526.2 | 405.6 |
| $1 \times 10^5$ | -1812.0 | -980.5 | 163.4 | 355.3 |
| $1 \times 10^6$ | -1357.0 | -988.0 | -387.9 | 264.4 |
| Pts | | | | |
| $1 \times 10^2$ | -3523.0 | -953.8 | 679.0 | 535.3 |
| $1 \times 10^3$ | -3178.0 | -992.0 | 606.0 | 511.2 |
| $1 \times 10^4$ | -2806.0 | -985.0 | 493.7 | 404.2 |
| $1 \times 10^5$ | -1816.0 | -980.6 | 152.4 | 354.2 |
| $1 \times 10^6$ | -1359.0 | -994.0 | -406.4 | 265.8 |

Table 3: Summary of $K_\alpha$ values per antigen type when different segment sizes are applied in ABS.

| Seg | Min | Mean | Max | Stdev |
|---|---|---|---|---|
| Nmap | | | | |
| 1 | -2248.0 | -966.2 | 539.3 | 387.7 |
| 10 | -1630.0 | -969.0 | 290.2 | 357.6 |
| $1 \times 10^2$ | 964.9 | -1065.0 | -27.1 | 310.6 |
| $1 \times 10^3$ | -1008.0 | -1049.0 | -796.3 | 155.0 |
| Firefox | | | | |
| 1 | -3445.0 | -1146.0 | 561.7 | 513.2 |
| 10 | -2153.0 | -1083.0 | 293.5 | 388.1 |
| $1 \times 10^2$ | -1080.0 | -1114.0 | 42.6 | 325.7 |
| $1 \times 10^3$ | -1066.0 | -1074.0 | -797.7 | 214.0 |
| Pts | | | | |
| 1 | -3445.0 | -1142.0 | 540.4 | 515.1 |
| 10 | -2227.0 | -1082.0 | 285.7 | 391.2 |
| $1 \times 10^2$ | -1082.0 | -1110.0 | -1.2 | 318.4 |
| $1 \times 10^3$ | -1065.0 | -1065.0 | -804.1 | 213.6 |

Table 4: Summary of $K_\alpha$ values per antigen type when different segment sizes are applied in TBS.

| | $1 \times 10^3$ | $1 \times 10^4$ | $1 \times 10^5$ | $1 \times 10^6$ |
|---|---|---|---|---|
| Nmap | | | | |
| $1 \times 10^2$ | 0.24 | 0.65 | 0.93 | 0.75 |
| $1 \times 10^3$ | – | 0.97 | 0.94 | 0.81 |
| $1 \times 10^4$ | – | – | 0.95 | 0.80 |
| $1 \times 10^5$ | – | – | – | 0.80 |
| Firefox | | | | |
| $1 \times 10^2$ | < 0.05 * | < 0.05 * | 0.57 | 0.74 |
| $1 \times 10^3$ | – | 0.50 | 0.68 | 0.94 |
| $1 \times 10^4$ | – | – | 0.88 | 0.97 |
| $1 \times 10^5$ | – | – | – | 0.92 |
| Pts | | | | |
| $1 \times 10^2$ | < 0.05 * | < 0.05 * | 0.39 | 0.60 |
| $1 \times 10^3$ | – | 0.56 | 0.72 | 0.98 |
| $1 \times 10^4$ | – | – | 0.89 | 0.91 |
| $1 \times 10^5$ | – | – | – | 0.87 |

Table 5: The p-value of two-sample two-sided t-tests for ABS ('*' indicates a significant difference).

| | 10 | $1 \times 10^2$ | $1 \times 10^3$ |
|---|---|---|---|
| Nmap | | | |
| 1 | 0.89 | 0.98 | 0.63 |
| 10 | – | 0.94 | 0.65 |
| $1 \times 10^2$ | – | – | 0.66 |
| Firefox | | | |
| 1 | < 0.05 | 0.17 | 0.51 |
| 10 | – | 0.94 | 0.88 |
| $1 \times 10^2$ | – | – | 0.91 |
| Pts | | | |
| 1 | < 0.05 | 0.20 | 0.52 |
| 10 | – | 0.99 | 0.88 |
| $1 \times 10^2$ | – | – | 0.89 |

Table 6: The p-value of two-sample two-sided t-tests for TBS ('*' indicates a significant difference).

H1 and H2. The comparisons are performed within each segmentation approach, by comparing the experimental result of one segment size with the experimental result of another. As shown in Table 5, in ABS changing segment size does not make any significant differences to the $K_\alpha$ values of Nmap, but it can make significant differences to the $K_\alpha$ values of Firefox or Pts. Therefore, H1 is rejected. Conversely, as shown in Table 6, changing segment size cannot cause any significant differences to $K_\alpha$ values of Nmap, but it can cause significant differences to $K_\alpha$ values of Firefox or Pts. As a result, H2 is rejected.

In order to test H3, the experimental results of segmentation approaches are compared with the result of the standard dDCA (non-segmentation). The comparisons are performed per antigen type for every segment size. As no randomness is involved in dDCA, the same sequence of processed information is analysed in both segmentation and non-segmentation approaches. The difference is that the standard dDCA produces only one set of $K_\alpha$ values of each antigen type, whereas systems with segmentation produce multiple sets (equal to the number of generated segments) of $K_\alpha$ values. Therefore, the $K_\alpha$ values produced by the standard dDCA are used as the true means in the one-sample one sided t-tests ($\alpha$ = 0.05) [3], to test whether the means of the $K_\alpha$ val- ues produced by segmentation approaches are significantly different. This can indicate whether applying segmentation can produce significantly different or better detection per- formance. The p-value of all tests are listed in Table 7. If the p-value is less than 0.05, for anomalous antigen types (intrusions), it implies that segmentation approaches pro- duce significant different and better results. Whereas, for normal antigen types (not intrusions), this indicates that segmentation approaches can produce significantly different but not necessarily better results. As shown in the table, when segment size is equal to $1 \times 10^2$, $1 \times 10^3$, or $1 \times 10^4$, ABS can produce significantly different and better results of Nmap and Pts. When segment size is $1 \times 10^5$ or $1 \times 10^6$, ABS can make significant differences to the results of Firefox, but not the results of Nmap or Pts. Conversely, TBS can make significant differences to the results of Firefox, but not the results of Nmap or Pts. In summary, segmentation approaches can make significant differences the results, and ABS can produce better performance on identifying intrusions while TBS cannot. Therefore, H3 is rejected.

Even though no direct statistical tests are performed to test H4, the statistical tests of other hypotheses can indicate whether it should be rejected or not. Firstly, changing segment size in TBS has less effect on the results than ABS, as shown in Table 5 and Table 6. Secondly, ABS can produce significantly different and better results on identifying intrusions, while TBS cannot. Therefore, changing segmentation approach from ABS to TBS can make differences to the DCA, and H4 is rejected.

As described above, applying segmentation makes significant differences to the results, since it has changed the way of analysing processed information. It can improve the results in terms of identifying intrusions, this applies to the ABS approach in particular. However, segmentation cannot improve the results on tolerating normal processes, this may be because of the 'innocent bystander effect' of the DCA [7]. This effect occurs when normal processes are highly active and appear at the same time as the anomalous processes, the algorithm could classify normal processes as intrusions (false alarms). This is an inherited issue of the dataset which has not been resolved. In addition, it appears that ABS performs better than TBS. It is possible that the number of sampled antigens is more vital to the analysis process than the processed time. As a result, ABS can always assure that each segment includes sufficient processed information, making it more effective for analysis. Moreover, changing segment size in both approaches can make significant differences to the results. This is because it produces the quantitative differences of processed information per segment that is analysed by the analysis process. Furthermore, changing segmentation from ABS to TBS can make a difference to the detection performance. This may result from the fact that ABS can ensure sufficient processed information is included within every segment for analysis. However, TBS cannot provide such assurance. As mentioned previously, some segments in TBS can contain few or even no sampled antigens, therefore nothing can be detected in such segments, as there is nothing to classify.

| Seg | Nmap | Firefox | Pts |
|---|---|---|---|
| | -966.21 | -1389.04 | -1005.66 |
| Antigen based segmentation | | | |
| $10^2$ | $< 0.05$ * | $< 0.05$ * | $< 0.05$ * |
| $10^3$ | $< 0.05$ * | $< 0.05$ * | $< 0.05$ * |
| $10^4$ | $< 0.05$ * | $< 0.05$ * | $< 0.05$ * |
| $10^5$ | 0.14 | $< 0.05$ * | 0.21 |
| $10^6$ | 0.41 | $< 0.05$ * | 0.44 |
| Time based segmentation | | | |
| 1 | 0.50 | $< 0.05$ * | 1 |
| 10 | 0.56 | $< 0.05$ * | 1 |
| $10^2$ | 0.49 | $< 0.05$ * | 0.95 |
| $10^3$ | 0.69 | $< 0.05$ * | 0.69 |

Table 7: The p-value of one-sample one-sided t-tests, true means are listed in the second row ('*' indicates a significant difference).

## 6. CONCLUSIONS AND FUTURE WORK

In this paper, we have shown that applying segmentation to the DCA makes significant differences to the results. In fact, the ABS approach can improve the results, in terms of identifying intrusions. In addition, segmentation enables the system to perform periodic analysis on the processed information presented by the DCs. As a result it can effectively improve detection speed without compromising detection accuracy. Therefore, segmentation is applicable to the DCA. Even though segmentation is not immune inspired, it can still make contribution to the field of AIS, as it can improve the system performance of the DCA. As a result, more effective intrusion detection systems can be developed by integrating segmentation with the DCA. This method is also applicable to other second generation AIS.

This is not yet real-time analysis, as the segmentation approaches still occur after a time delay. The actual situations occur during detection always change in a dynamic fashion. Therefore, in order to perform real-time analysis, an approach that can deal with online dynamics is required. Such an approach should be able to adapt and evolve during detection, so that it can deal with new situations that have not been previously seen. This leads to the future work of dynamic segmentation.

To continue this research, dynamic segmentation is to be explored to develop more effective intrusion detection systems. Moreover, as the ultimate goal is to develop a real-time system for the purpose of intrusion detection, it is necessary to use formal techniques available in the area of real-time systems. For example, we can use Duration Calculus [18] to specify the real-time system demanded, and Timed Automata [2] and PLC Automata [5] to implement and verify such system.